\documentclass[runningheads,envcountsame, a4paper]{llncs}
\usepackage[pdftex]{graphicx}
\usepackage{epstopdf}
\usepackage{xcolor}
\usepackage{lipsum}
\usepackage{authblk}
\usepackage{amsmath,amssymb,mathtools,bm,etoolbox}
\usepackage{algorithmic}
\usepackage[ruled,vlined]{algorithm2e}
\usepackage[super]{nth}
\usepackage[section]{placeins}
\usepackage[hyphens]{url}
\usepackage{booktabs}
\usepackage{siunitx}
\usepackage{multirow}
\usepackage{subfigure}
\usepackage{rotating}
\usepackage{mwe}
\newcolumntype{L}{>{\centering\arraybackslash}m{2cm}}
\usepackage[normalem]{ulem}
\usepackage[misc]{ifsym}

\begin{document}

\title{Few-Shot Microscopy Image Cell Segmentation}
\titlerunning{Few-Shot Microscopy Image Cell Segmentation}
\toctitle{Few-Shot Microscopy Image Cell Segmentation}

\author{Youssef Dawoud (\Letter) \inst{1}\and
Julia Hornauer  \inst{1}\and
Gustavo Carneiro \inst{2}\and
\\Vasileios Belagiannis \inst{1}}

\authorrunning{Y.Dawoud, J.Hornauer, G.Carneiro, and V.Belagiannis}

\institute{Universit\"at Ulm, Ulm, Germany  \and The University of Adelaide, Adelaide, Australia \\ \email{firstname.lastname@\{uni-ulm.de, adelaide.edu.au\}}}

\maketitle    \setcounter{footnote}{0}         
\begin{abstract}

Automatic cell segmentation in microscopy images works well with the support of deep neural networks trained with full supervision. Collecting and annotating images, though, is not a sustainable solution for every new microscopy database and cell type. Instead, we assume that we can access a plethora of annotated image data sets from different domains (sources) and a limited number of annotated image data sets from the domain of interest (target), where each domain denotes not only different image appearance but also a different type of cell segmentation problem. We pose this problem as meta-learning where the goal is to learn a generic and adaptable few-shot learning model from the available source domain data sets and cell segmentation tasks. The model can be afterwards fine-tuned on the few annotated images of the target domain that contains different image appearance and different cell type. In our meta-learning training, we propose the combination of three objective functions to segment the cells, move the segmentation results away from the classification boundary using cross-domain tasks, and learn an invariant representation between tasks of the source domains. Our experiments on five public databases show promising results from 1- to 10-shot meta-learning using standard segmentation neural network architectures.

\end{abstract}

\section{Introduction}

Microscopy image analysis involves many procedures including cell counting, detection and segmentation~\cite{xing2016robust}. Cell segmentation is particularly important for studying the cell morphology to identify the shape, structure, and cell size. Manually segmenting cells from microscopy images is a time-consuming and costly process. For that reason, many methods have been developed to automate the process of cell segmentation, as well as counting and detection.

Although reliable cell segmentation algorithms exist for more than a decade~\cite{wahlby2004combining,xing2016robust}, only recently they have shown good generalization with the support of deep neural networks~\cite{ciresan2012deep,xie2018microscopy}. Current approaches in cell segmentation deliver promising results based on encoder - decoder network architectures trained with supervised learning~\cite{xie2018microscopy}. However, collecting and annotating large amounts of images is practically inviable for every new microscopy problem. Furthermore, new problems may contain different types of cells, where a segmentation model, pre-trained on different data sets, may not deliver good performance. 

\begin{figure}[!htbp]
\centering

    \rotatebox[origin=l]{90}{TNBC Database}
	\includegraphics[width=3.8cm]{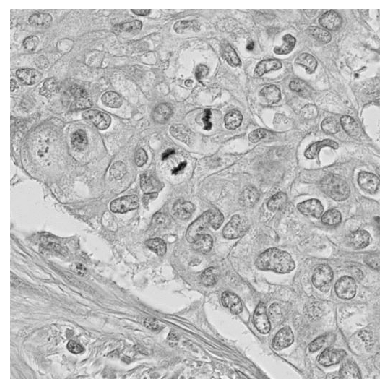}
	\includegraphics[width=3.8cm]{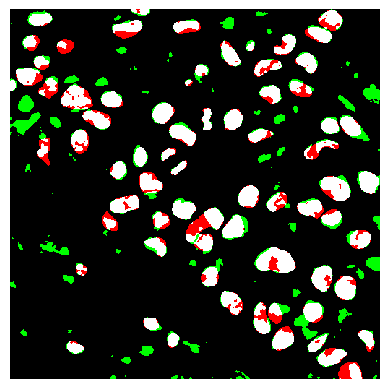}
	\includegraphics[width=3.8cm]{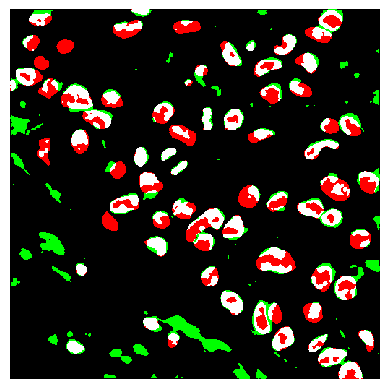}
	
	\rotatebox[origin=l]{90}{ssTEM Database}
	\includegraphics[width=3.8cm]{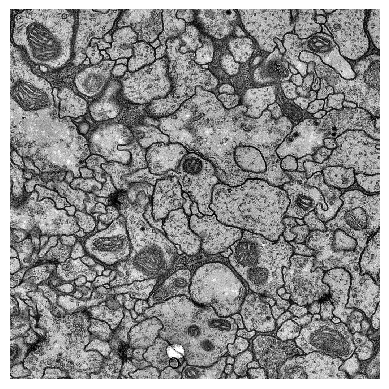}
	\includegraphics[width=3.8cm]{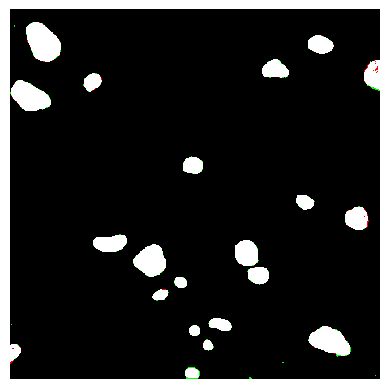}
	\includegraphics[width=3.8cm]{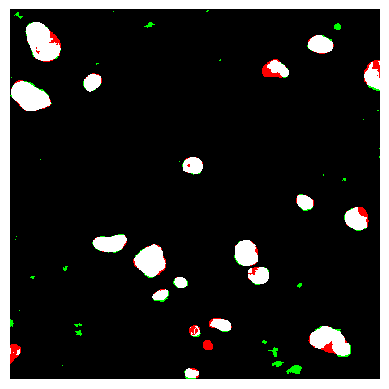}
	
	\rotatebox[origin=l]{90}{EM Database}
	\includegraphics[width=3.8cm]{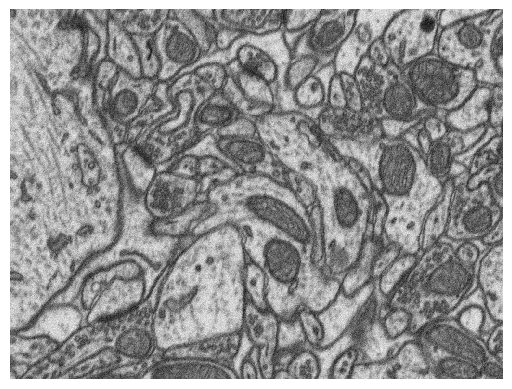}
	\includegraphics[width=3.8cm]{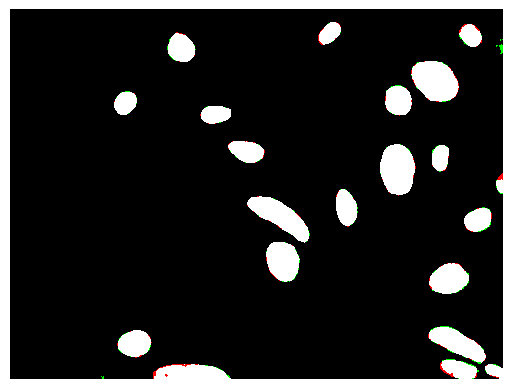}
	\includegraphics[width=3.8cm]{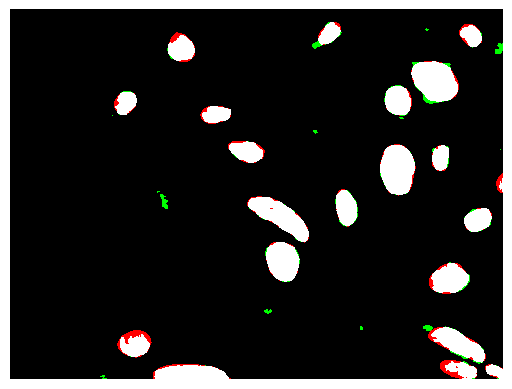}

	\rotatebox[origin=l]{90}{B5 Database}
	\includegraphics[width=3.8cm]{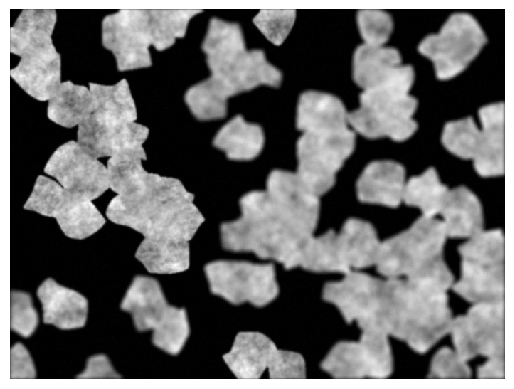}
	\includegraphics[width=3.8cm]{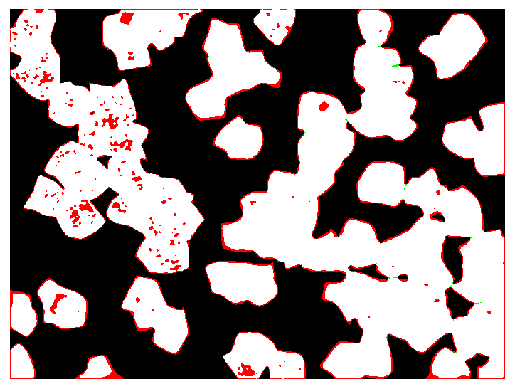}
	\includegraphics[width=3.8cm]{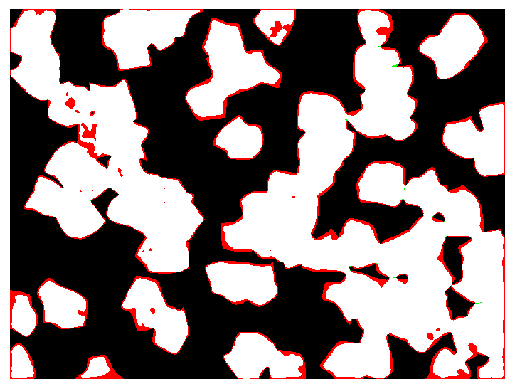}
	
	\rotatebox[origin=l]{90}{B39 Database}
	\subfigure[c][Input]{\includegraphics[width=3.8cm]{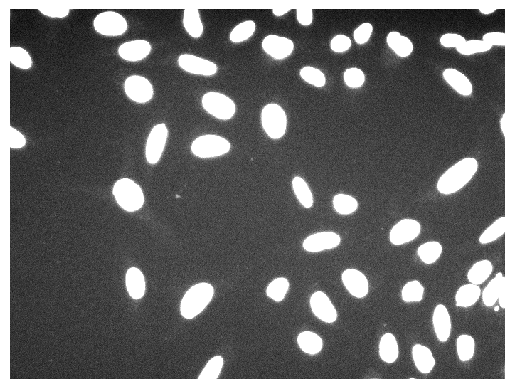}}
	\subfigure[c][Ours]{\includegraphics[width=3.8cm]{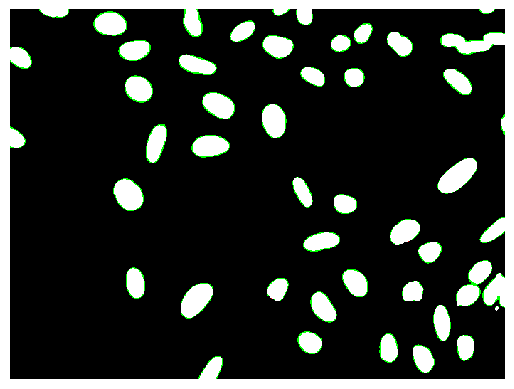}}
    \subfigure[c][Transfer Learning]{\includegraphics[width=3.8cm]{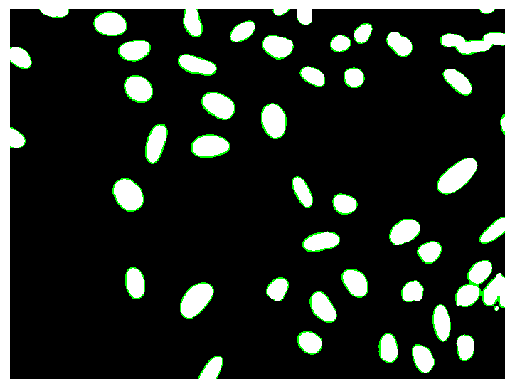}}

	\caption{Visual Result. We visually compare our approach to transfer learning using the U-Net architecture. Ours refers to meta-training with all objectives, namely $\mathcal{L}_{BCE} + \mathcal{L}_{ER} + \mathcal{L}_{D}$. The red color corresponds to false positive, the green color to false negative, the black color to true negative, and the white color to true positive. Best viewed in color.}
	\label{fig:VisualComp1}
\end{figure}

To address this limitation, methods based on domain generalization~\cite{dou2019domain,li2018learning}, domain adaptation~\cite{tzeng2017adversarial} and few-shot learning~\cite{ravi2016optimization} have been developed. In these approaches, it is generally assumed that we have access to a collection of annotated images from different domains (source domains), no access to the target domain (domain generalisation), access to a large data set of un-annotated images (domain adaptation), or access to a limited number (less than 10) of annotated images from the target domain (few-shot learning). Domain generalisation and adaptation generally involve the same tasks in the source and target domains, such as the same cell type segmentation problem with images coming from different sites and having different appearances.
However, our setup is different because we consider that the source and target domains consist of different types of cell segmentation problems. Each domain consists of different cell types such as mitochondria and nuclei. In this way, we form a typical real-life scenario where a variety of microscopy images, containing different cell structures, are leveraged from various resources to learn a cell segmentation model. Therefore, the challenge in our setup is to cope with different image and cell appearances, as well as different types of cell for each domain, as illustrated in Fig.~\ref{fig:VisualComp1}. In such setup, we argue that few-shot learning is more appropriate, where we aim to learn a generic and adaptable few-shot learning model from the available source domain data sets. This model can be afterwards fine-tuned on the few annotated images of the new target domain. Such problem can be formulated as an optimization-based meta-learning approach~\cite{dong2018few,finn2017model,nichol2018first}.

In this paper, we present a new few-shot meta-learning cell segmentation model. For meta-training the model, we propose the combination of three loss functions to 1. impose pixel-level segmentation supervision, 2. move the segmentation predictions away from the classification boundary using cross-domain tasks and 3. learn an invariant representation between tasks. In our evaluations on five microscopy data sets, we demonstrate promising results compared to the related work on settings from 1- to 10-shot learning by employing standard network architectures~\cite{ronneberger2015u,xie2018microscopy}. To the best of our knowledge, this is the first work to introduce few-shot task generalisation using meta-learning and to apply few-shot learning to microscopy image cell segmentation.

\section{Related Work} \label{relatedwork}

Cell segmentation is a well-established problem~\cite{xing2016robust} that is often combined with cell counting~\cite{arteta2016counting,dijkstra2018centroidnet} and classification~\cite{zhang2012classifying}. We discuss the prior work that is relevant to our approach, as well as the related work on few-shot learning. 
\subsection{Cell Segmentation}

Automatic cell detection and counting in microscopy images has been earlier studied with the support of image processing and computer vision techniques~\cite{faustino2009automatic,lu2015improved,wahlby2004combining}. In past few years, advances in deep neural networks have changed the type of approaches developed for microscopy images. In particular, fully convolutional neural networks (FCNs)~\cite{long2015fully} allow to make predictions on the same or similar spatial resolution of the input image. FCN approaches have been widely adapted in medical imaging with applications to nuclei segmentation~\cite{naylor2017nuclei}, brain tumor segmentation from magnetic resonance imaging (MRI)~\cite{de2015deep} and of course segmentation in microscopy images~\cite{ciresan2012deep}. For instance, histology image segmentation relies on FCNs~\cite{cirecsan2013mitosis} to perform mitosis detection. Among the FCN models, U-Net~\cite{ronneberger2015u} is a very popular architecture. It has been developed for segmenting neuronal structures in electron microscopy images, but it is presently used for any kind of medical imaging that demands spatial predictions. Similarly, the fully convolutional regression network (FCRN)~\cite{xie2018microscopy} is another encoder - decoder architecture for cell segmentation and counting in microscopy images. In the evaluation, we consider both U-Net and FCRN architectures.

The main difference between our approach and existing cell segmentation approaches lies in the training algorithm. While the aforementioned approaches deliver promising results on the examined data sets, they all require a large amount of annotated data and a fully supervised training process. In this work, we address the problem as a more realistic few-shot learning problem, where we have access to relatively large data sets of different domain and of several types of cell segmentation problems, but the data set for the target segmentation domain is limited with just a handful of training samples. We present an approach to reach promising segmentation performance regardless of the small target segmentation data set.

\subsection{Few-shot Learning}

Few-shot learning deals with making use of existing knowledge, in terms of data and annotation, to build a generic model that can be adapted to different (but related) target problems with limited training data. Although there are classic approaches from the past~\cite{mensink2012metric,rohrbach2013transfer}, deep neural networks and meta-learning have significantly contributed to improve the state of the art of few-shot learning. In meta-learning, the goal is to train a model to learn from a number of tasks using a limited number of training samples per task~\cite{schmidhuber1992learning}. Meta-learning, in general, consists of a meta-training phase where multiple tasks adapt a base learner to work with different problems (where each task uses a small training set), then the classifiers of those multiple tasks are pooled together to update the base learner. After the meta-training process converges, the model of the base learner is fine-tuned to a limited number of annotated data sampled from the target problem. Meta-learning approaches can be categorized in metric-learning~\cite{snell2017prototypical,vinyals2016matching}, model-~\cite{clavera2018model,munkhdalai2017meta} and optimization-based learning~\cite{finn2017model,nichol2018first,ravi2016optimization}. The optimization meta-learning approaches function with gradient-based learning, which is less complex to implement and computationally efficient. For that reason, we rely on Reptile~\cite{nichol2018first}, an optimization-based approach, to develop our algorithm.

\section{Few-shot Cell Segmentation}
\label{cellsegmentation}

In this section, we define the few-shot segmentation problem. Afterwards, we pose cell segmentation from microscopy images as few-shot meta-learning approach.

\subsection{Problem Definition}\label{problemDef}

Let $\mathcal{S} = \{ \mathcal{S}_{1}, \mathcal{S}_{2},\dots, \mathcal{S}_{|\mathcal{S}|}\}$ be a collection of $|\mathcal{S}|$ microscopy cell data sets. Each data set $\mathcal{S}_{m}$ is defined as $\mathcal{S}_{m}=\{(\mathbf{x},\mathbf{y})_{k}\}_{k=1}^{|\mathcal{S}_{m}|}$, where $(\mathbf{x},\mathbf{y})_{k}$ is a pair of the microscopy cell image $\mathbf{x}:\Omega \rightarrow \mathbb{R}$ ($\Omega$ denotes the image lattice with dimensions $H \times W$) and the respective ground-truth segmentation mask $\mathbf{y}:\Omega \rightarrow \{0,1\}$; and $|\mathcal{S}_{m}|$ is the number of samples in $\mathcal{S}_{m}$. Note that each data set corresponds to a different task and domain, each representing a new type of image and segmentation task. All data sets in $\mathcal{S}$ compose the source data sets. We further assume another data set, which we refer to as target, defined by $\mathcal{T}=\{(\mathbf{x},\mathbf{y})_{k}\}_{k=1}^{|\mathcal{T}|}$, where the number of training images with ground-truth segmentation masks is limited, e.g.~between 1 and 10 training images and segmentation mask pairs. Also, we assume that the target data set comes from a different task and domain.

Our goal is to perform cell segmentation in the images belonging to the target data set $\mathcal{T}$ through a segmentation model $\mathbf{y}=f(\mathbf{x};\theta)$, where $\theta$ denotes the model parameter (i.e.~the weights of the deep neural network). However, the limited number of annotated training images prohibits us from learning a typical fully-supervised data-driven model. This is a common problem in real-life when working with cell segmentation problems, where annotating new data sets, i.e. the target data sets, does not represent a sustainable solution. To address this limitation, we propose to learn a generic and adaptable model from the source data sets in $\mathcal{S}$, which is then fine-tuned on the limited annotated images of the target data set $\mathcal{T}$.

\subsection{Few-shot Meta-Learning Approach}

 We propose to learn a generic and adaptable few-shot learning model with gradient-based meta-learning~\cite{finn2017model,nichol2018first}. The learning process is characterized by a sequence of episodes, where an episode consists of the meta-training and meta-update steps. In meta-training, the model parameter $\theta$ (this is known as the meta-parameter) initialises the segmentation model, with $\theta_m=\theta$ in $\mathbf{y} = f(\mathbf{x};\theta_m)$ (defined in Sec.~\ref{problemDef}) for each task $m \in \{1,...,|\mathcal{S}|\}$, where each task is modeled with a training set $\tilde{\mathcal{S}}_{m} \subset \mathcal{S}_m$.  Next, the model meta-parameter $\theta$ is meta-updated from the learned task parameters $\{ \theta_m \}_{m=1}^{|\mathcal{S}|}$.

 The segmentation model that uses the meta parameter $\theta$, defined as $\mathbf{y}=f(\mathbf{x};\theta)$, is denoted by the base learner.  To train this base learner, we propose three objective functions that account for segmentation supervision, and two regularisation terms. In our implementation, we rely on the Reptile algorithm~\cite{nichol2018first} to develop our approach because of its simplicity and efficiency. Our approach is described in Algorithm~\ref{algo}. Next, we present each part of the approach.
 
 \begin{algorithm}[!htbp]
\caption{Gradient-Based Meta-Learning for Cell Segmentation}\label{algo}
  \begin{algorithmic}[1]
  \STATE Input: source domain data sets $\mathcal{S} = \{ \mathcal{S}_{1}, \mathcal{S}_{2},\dots, \mathcal{S}_{|\mathcal{S}|}\}$, with $\mathcal{S}_{m}=\{(\mathbf{x},\mathbf{y})_{k}\}_{k=1}^{|\mathcal{S}_{m}|}$
  \STATE Input: target domain data set $\mathcal{T}$, with $\mathcal{T}=\{(\mathbf{x},\mathbf{y})_{k}\}_{k=1}^{|\mathcal{T}|}$
  \STATE Initialize $\theta,\alpha,\beta$, where $0<\alpha,\beta<1$
  \FOR{iteration = $1,2,\dots$}
  \FOR{$m$ = $1,2,\dots, |\mathcal{S}|$}
      \STATE Sample $K$ image and segmentation samples to form \\ $\tilde{\mathcal{S}_m} = \{(\mathbf{x},\mathbf{y})_k\}_{k=1}^{K} \subset \mathcal{S}_{m}$ from current task $\mathcal{S}_{m}$ 
      \STATE Randomly choose a different task $n \in \{1,...,|\mathcal{S}|\}$ such that $n \neq m$
      \STATE Randomly choose a different task $p \in \{1,...,|\mathcal{S}|\}$ such that $p \neq m$
      \STATE Sample $K$ image samples from task $n$ to form $\tilde{\mathcal{S}_n} =  \{(\mathbf{x})_k\}_{k=1}^{K} \subset \mathcal{S}_{n}$ 
      \STATE Sample $K$ image samples from task $p$ to form $\tilde{\mathcal{S}_p} =  \{(\mathbf{x})_k\}_{k=1}^{K} \subset \mathcal{S}_{p}$ 
     \STATE Compute gradient $\theta^{\prime}_{m} = g(\mathcal{L}_{BCE}(\theta,\tilde{\mathcal{S}_m})+\alpha\mathcal{L}_{ER}(\theta,\tilde{\mathcal{S}_n})+\beta\mathcal{L}_{D}(\theta,\tilde{\mathcal{S}_m},\tilde{\mathcal{S}_p}))$
  \ENDFOR
  \STATE Meta-update $\theta \leftarrow \theta + \epsilon \frac{1}{|\mathcal{S}|}\sum_{m=1}^{|\mathcal{S}|}(\theta^{\prime}_{m} - \theta)$
 \ENDFOR
 \STATE Produce few-shot training set from target by sampling $K$ samples to form $\tilde{\mathcal{T}}= \{(\mathbf{x},\mathbf{y})_k\}_{k=1}^{K} \subset \mathcal{T}$
 \STATE Fine-tune $f(\mathbf{x};\theta)$ with $\tilde{\mathcal{T}}$ using $\mathcal{L}_{BCE}(\theta,\tilde{\mathcal{T}})$
 \STATE Test $f(\mathbf{x};\theta)$ with testing set $\hat{\mathcal{T}} = \mathcal{T} \setminus \tilde{\mathcal{T}}$.
 \end{algorithmic}
 \end{algorithm}

\subsection{Meta-Learning Algorithm} 

During meta-training, i.e.~lines 5 to 12 in Algorithm~\ref{algo}, $|\mathcal{S}|$ tasks are generated by sampling $K$ images from each source domain in $\mathcal{S}$. Hence, a task is represented by a subset of $K$ images and segmentation maps from $\mathcal{S}_m$. In our experiments, we work from $K\in\{1,...,10\}$-shot learning problems. After sampling a task, the baser learner is trained with the three objective functions. The first objective function consists of the standard  binary cross entropy loss $\mathcal{L}_{BCE}(.)$ that uses the images and segmentation masks of the sampled task. The second loss function is based on the entropy regularization $\mathcal{L}_{ER}(.)$~\cite{grandvalet2005semi} that moves the segmentation results away from the classification boundary using a task $n \neq m$ without the segmentation maps from task $n$ -- such regularisation makes the segmentation results for task $n$ more confident, i.e.~more binary like. The third objective loss function consists of extracting an invariant representation between tasks~\cite{dou2019domain} by enforcing the learning of a common feature representation across different tasks with the knowledge distillation loss $\mathcal{L}_{D}(.)$~\cite{hinton2015distilling}. The use of the entropy regularisation and knowledge distillation losses in few-shot meta-learning represent the main technical contribution of our approach. The learned task parameters $\theta^{\prime}_m$ are optimized using the three objective functions above during meta-training, as follows:
\begin{equation}\label{meta-train-eq}
    \theta^{\prime}_m = \arg \min_{\theta} \mathbb{E}_{\tilde{\mathcal{S}}_{m},\tilde{\mathcal{S}}_n,\tilde{\mathcal{S}}_p} [\mathcal{L}(\theta,\tilde{\mathcal{S}}_{m},\tilde{\mathcal{S}}_n,\tilde{\mathcal{S}}_p)],
\end{equation}
where the loss $\mathcal{L}(\theta,\tilde{\mathcal{S}}_{m},\tilde{\mathcal{S}}_n,\tilde{\mathcal{S}}_p) = \mathcal{L}_{BCE}(\theta,\tilde{\mathcal{S}}_{m}) + \alpha \mathcal{L}_{ER}(\theta,\tilde{\mathcal{S}}_{n}) + \beta \mathcal{L}_{D}(\theta,\tilde{\mathcal{S}}_{m},\tilde{\mathcal{S}}_p)$ is a combination of the three objectives that depend on $K$-shot source domain training sets $\tilde{\mathcal{S}}_{\{m,n,p\}} \subset \mathcal{S}_{\{m,n,p\}}$ (with $m \neq n$, $m \neq p$ and $\{m,n,p\}\subset\{1,...,|\mathcal{S}| \}$) and $|\tilde{\mathcal{S}}_{\{m,n,p\}}|=K$. In Sec.~\ref{objectives}, we present in detail the objective functions. Finally, the parameters of the base learner are learned with stochastic gradient descent and back-propagation. 

At last, the meta-update step takes place after iterating over $|\mathcal{S}|$ tasks during meta-training. The meta-update, i.e. line 11 in Algorithm~\ref{algo}, updates the model parameter $\theta$ using the following rule:
\begin{equation}\label{meta-update-eq}
    \theta \leftarrow \theta + \epsilon \frac{1}{|\mathcal{S}|}\sum_{m=1}^{|\mathcal{S}|}(\theta^{\prime}_{m} - \theta),
\end{equation}
where $\epsilon$ is the step-size for the parameter update. The episodic training takes place based on the meta-training and meta-update steps until convergence.

\subsection{Task Objective Functions}
\label{objectives}

We design three objective functions for the meta-training stage to update the parameters of the base learner. 
 
\subsubsection{Binary Cross Entropy Loss}\label{CEntropyLoss}

Every sampled task includes pairs of input image and segmentation mask. We rely on the pixel-wise binary cross entropy (BCE) as the main objective to learn predicting the ground-truth mask. In addition, we weight the pixel contribution since we often observe unbalanced pixel ratio between the foreground and background pixels. Given the foreground probability prediction $\mathbf{y}^{\prime} = f(\mathbf{x};\theta)$ of the input image $\mathbf{x}$ and the segmentation mask $\mathbf{y}$ that belong to the $K$-shot training set $\tilde{\mathcal{S}}_m$ for task $m$, the pixel-wise BCE loss is given by:

\begin{equation}\label{bce_seg}
    \begin{aligned}
     \mathcal{L}_{BCE} (\theta,\tilde{\mathcal{S}}_m) &=
         - \frac{1}{|\tilde{\mathcal{S}}_m|}\sum_{ (\mathbf{x},\mathbf{y}) \in \tilde{\mathcal{S}}_m } \sum_{\omega \in \Omega }^{ }\left[\mathbf{w} \mathbf{y}(\omega)\log( \mathbf{y}^{\prime}(\omega) ) +   \right. \\
         &\left. (1-\mathbf{y}(\omega)) \log(1-\mathbf{y}^{\prime}(\omega)) 
         \right] ,
\end{aligned}
\end{equation}

where $\omega \in \Omega$ denotes to the spatial pixel position in the image lattice $\Omega$ and $\mathbf{w}$ is the weighting factor of the the foreground class probability $\mathbf{y}(\omega)$ which equals to the ratio of background to foreground classes in $\mathbf{y}$. This is the standard loss function for segmentation-related problems, which we also employ for our binary problem~\cite{chen2017deeplab}.

\subsubsection{Entropy Regularization}\label{EntropyReg}
The BCE loss can easily result in over-fitting the base-learner to the task $m$. We propose to use Shannon's entropy regularization on a different task loss to prevent this behavior. More specifically, while minimizing the BCE loss on a sampled task of one source domain, e.g.~task $m$, we sample a second task $n \neq m$ from a different domain, e.g.~task $n$; and seek to minimize Shannon's entropy loss for task $n$ without using that segmentation masks of that task. As a result, while minimizing the BCE loss for task $m$, we are also aiming to make confident predictions for task $n$ by minimizing Shannon's entropy. The regularization is defined as:
\begin{equation}\label{entropy_req}
    \mathcal{L}_{ER} (\theta,\tilde{\mathcal{S}}_n) = \frac{1}{|\tilde{\mathcal{S}}_n|}\sum_{ (\mathbf{x}) \in \tilde{\mathcal{S}}_n } \sum_{\omega \in \Omega }^{ } \begin{bmatrix} \mathcal{H} (\mathbf{y}^{\prime}(\omega))\end{bmatrix},
\end{equation}
where Shannon's entropy is $\mathcal{H} (\mathbf{y}^{\prime}(\omega)) = - \mathbf{y}^{\prime}(\omega)\log(\mathbf{y}^{\prime}(\omega))$ for the foreground pixel probability $\mathbf{y}^{\prime}(\omega)$, and $\mathbf{y}^{\prime} = f(\mathbf{x};\theta)$. Our motivation for the entropy regularizer originates from the field of semi-supervised learning~\cite{grandvalet2005semi}. The same loss has been recently applied to few-shot classification~\cite{dhillon2019baseline}. 

\subsubsection{Distillation Loss}\label{DistillLoss}

Although we deal with different source domains and cell segmentation problems, we can argue that microscopy images of cells must have common features in terms of cell morphology (shape and texture) regardless of the cell type. By learning from as many source data sets as possible, we aim to have a common representation that addresses the target data set to some extent. We explore this idea during meta-training by constraining the base learner to learn a common representation between different source data sets. To that end, two tasks from two source data sets $m,p$ ($m \neq p$, $m,p \in \{1,...,|\mathcal{S}|$) are sampled in every meta-training iteration. Then, the Euclidean distance between the representations of two images (one from each task) from the $l^{th}$ layer of the segmentation model is minimised. This idea is common in knowledge distillation~\cite{hinton2015distilling} and network compression~\cite{belagiannis2018} where the student network learns to predict the same representation as the teacher network. Recently, the idea of distillation between tasks has been also used for few-shot learning~\cite{dou2019domain}. We sample an image $\mathbf{x}^{(m)}$ from data set $m$ and $\mathbf{x}^{(p)}$ from the data set $p$. Then, we define the distillation loss as:
\begin{equation}\label{kd_eq}
    \mathcal{L}_{D} (\theta,\tilde{\mathcal{S}}_m,\tilde{\mathcal{S}}_p) = \frac{1}{|\tilde{\mathcal{S}}_m||\tilde{\mathcal{S}}_p|}\sum_{ (\mathbf{x}^{(m)}) \in \tilde{\mathcal{S}}_m } \sum_{ (\mathbf{x}^{(p)}) \in \tilde{\mathcal{S}}_p } \begin{bmatrix} f^{l} (\mathbf{x}^{(m)};\theta) - f^{l} (\mathbf{x}^{(p)};\theta) \end{bmatrix}^2,
\end{equation}
where $f^{l}(\mathbf{x}^{(m)};\theta)$ and $f^{l}(\mathbf{x}^{(p)};\theta)$ correspond to the $l$-th layer activation maps of the base learner for  images $\mathbf{x}^{(m)}$ and $\mathbf{x}^{(p)}$, respectively. Furthermore, the $l^{th}$ layer feature of the base learner representation is the latent code of an encoder - decoder architecture~\cite{ronneberger2015u,xie2018microscopy}.

\subsection{Fine-Tuning}

After the meta-learning process is finished, the segmentation model is fine-tuned using a $K$-shot subset of the target data set, denoted by $\tilde{\mathcal{T}} \subset \mathcal{T}$, with $|\tilde{\mathcal{T}}|=K$ (see line 16 in Algorithm~\ref{algo}). This fine-tuning process is achieved with the following optimisation:
\begin{equation}
    \theta^* = \arg \min_{\theta} [\mathcal{L}_{BCE}(\theta,\tilde{\mathcal{T}})].
\end{equation}
Here, we only need the binary cross entropy loss $\mathcal{L}_{BCE}$ for fine-tuning. We also rely on Adam optimization and back-propagation for fine-tuning, though we use different hyper-parameters as we later report in the implementation details (Sec.~\ref{implem_details}). At the end, our model with the updated parameters $\theta^*$ is evaluated on the target test set $\hat{\mathcal{T}} = \mathcal{T} \setminus \tilde{\mathcal{T}}$ (see~line 17 in Algorithm~\ref{algo}).

\section{Experiments}
\label{experiments}
We evaluate our approach on five microscopy image data sets using two standard encoder - decoder network architectures for image segmentation~\cite{ronneberger2015u,xie2018microscopy}. All evaluations take place for 1-, 3-, 5-, 7- and 10-shot learning, where we also compare with transfer-learning. In transfer learning the model is trained on all available source data sets and then fine-tuned on the few shots available from the target data set. This work is the first to explore few-shot microscopy image cell segmentation, so we propose the assessment protocol and data sets.

\subsection{Implementation Details}\label{implem_details}
We implement two encoder - decoder network architectures. The first architecture is the the fully convolutional regression network (FCRN) from~\cite{xie2018microscopy}. We moderately modify the decoder by using transposed convolutions instead of bi-linear up-sampling as we observed better performance. We also rely on sigmoid activation functions for making binary predictions, instead of the heat-map predictions because of the better performance too. The second architecture is the well-established U-Net~\cite{ronneberger2015u}. We empirically adapted the original architecture to a lightweight variant, where the number of layers are reduced from 23 to 12. We trained them in meta-training with Adam optimizer with learning rate 0.001 and weight decay of 0.0005. In meta-learning, we set $\epsilon$ from Eq.~\eqref{meta-update-eq} to 1.0. Both networks contain batch-normalization. However, batch-normalization is known to face challenges in gradient-based meta-learning due to the task-based learning~\cite{bronskill2020tasknorm}. The problem is on learning global scale and bias parameters of batch-normalization based on the tasks. For that reason, we only make use of the mean and variance during meta-training. On fine-tuning, we observed that the scale and bias parameters can be easily learned for the FCRN architecture. For U-Net, we do not rely on the scale and bias parameters for meta-learning. To allow a fair comparison with transfer-learning, we follow the same learning protocol of batch normalization parameters. During fine-tuning, we rely on the Adam optimizer with 20 epochs, learning rate 0.0001, and weight decay 0.0005. The same parameters are used for transfer learning. Our implementation and evaluation protocol is publicly available\footnote{https://github.com/Yussef93/FewShotCellSegmentation}.

\begin{figure}[!ht]
	\centering
	\subfigure[c][Input]{\includegraphics[height=1.75cm]{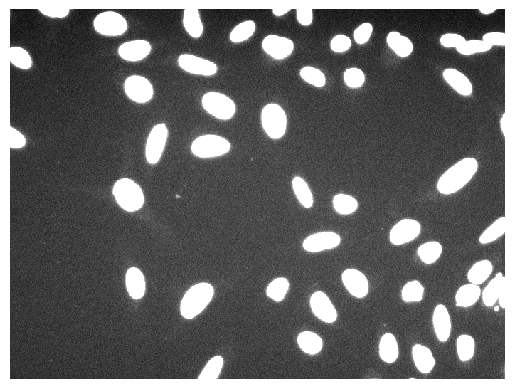}}
	\subfigure[c][$\mathcal{L}_{BCE}$]{\includegraphics[height=1.75cm]{_U-net_selection_2_overlap_batchidx_0_index_12.png}}
	\subfigure[c][$\mathcal{L}_{BCE} + \mathcal{L}_{ER}$]{\includegraphics[height=1.75cm]{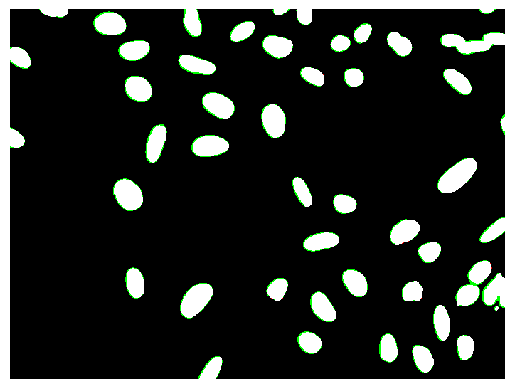}}
	\subfigure[c][$\mathcal{L}_{BCE} + \mathcal{L}_{D}$]{\includegraphics[height=1.75cm]{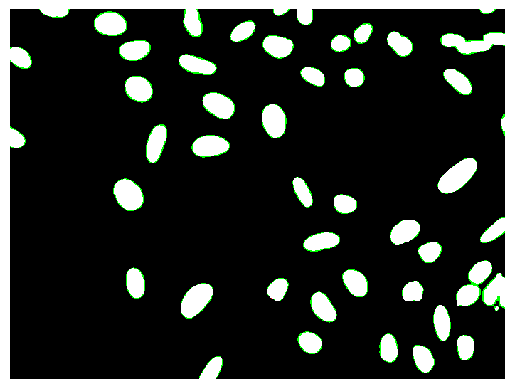}}
	\subfigure[c][Complete]{\includegraphics[height=1.75cm]{Regularizer_distillation_entropy__U-net_selection_2_overlap_batchidx_0_index_12}}
	
	\caption{Results of Our Objectives: We visually compare the effect of our objective functions in meta-training. The complete figure (e) refers to meta-training with all objectives, namely $\mathcal{L}_{BCE} + \mathcal{L}_{ER} + \mathcal{L}_{D}$. The green color corresponds to false negative, the white color to true positive and black to the true negative. Best viewed in color.}
	\label{fig:ModelCOmponents}
\end{figure}

\subsection{Microscopy Image Databases}

We selected five data sets of different cell domains, i.e. $|\mathcal{S}|=5$. First, we rely on the Broad Bioimage Benchmark Collection (BBBC), which is a collection of various microscopy cell image data sets~\cite{lehmussola2007computational}. We use BBBC005 and BBBC039 from BBBC, which we refer to as B5 and B39. B5 contains 1200 fluorescent synthetic stain cells, while B39 has 200 fluorescent nuclei cells. Second, the Serial Section Transmission Electron Microscopy (ssTEM)~\cite{Gerhard2013} database has 165 images of mitochondria cells in neural tissues. Next, the Electron Microscopy (EM) data set~\cite{lucchi2013learning} contains 165 electron microscopy images of mitochondria and synapses cells. Finally, the Triple Negative Breast Cancer (TNBC) database~\cite{naylor2018segmentation} has 50 histology images of breast biopsy. We summarize the cell type, image resolution and number of images for data sets in Table~\ref{5data sets}. Moreover, we crop the training images to $256\times256$ pixels, while during testing, the images are used in full resolution.

\begin{table}[!htbp]
\centering
\caption{Microscopy Image Databases. We present the details of the five data sets upon which we conduct our experiments. }
\label{tab:data set}
\begin{tabular}{|l|c|c|c|c|c|}
\hline
\multicolumn{1}{|l|}{data set} & B5~\cite{lehmussola2007computational} & B39~\cite{lehmussola2007computational} & ssTEM~\cite{Gerhard2013} & EM~\cite{lucchi2013learning} & TNBC~\cite{naylor2018segmentation} \\ \hline
Cell Type                     & synthetic stain   &nuclei     & mitochondria      & mitochondria    & nuclei      \\ \hline
Resolution                    &  $696\times520$   &$696\times520$     &  $1024\times1024$     &$768\times1024$      & $512\times512$      \\ \hline
\# of Samples                          &1200   &200     &20       &165      &50       \\ \hline
\end{tabular}\label{5data sets}
\end{table}

\subsection{Assessment Protocol}\label{protocol_eval}

We rely on the mean intersection over union (IoU) as the evaluation metric for all experiments. This is a standard performance measure in image segmentation~\cite{dong2018few,long2015fully}. We conduct a leave-one-dataset-out cross-validation~\cite{dou2019domain}, which is a common domain generalization experimental setup. This means that we use the four microscopy image data sets for meta-training and the remaining unseen data set for fine-tuning. From the remaining data set, we randomly select $K$ samples for fine-tuning. Since the selection of the $K$-shot images can vary significantly the final result, we repeat the random selection of the $K$-shot samples ten times and report the mean and standard deviation over these ten experiments. The same evaluation is performed for 1-,3-,5-,7- and 10-shot learning. Similarly, we evaluate the transfer learning approach as well.

\subsubsection{Objectives Analysis:} We examine the impact of each objective, as presented in Sec.~\ref{objectives}, to the final segmentation result. In particular, we first meta-train our model only with the binary cross entropy $\mathcal{L}_{BCE}$ from Eq.~\ref{bce_seg}. Second, we meta-train the binary cross entropy $\mathcal{L}_{BCE}$ jointly with entropy regularization $\mathcal{L}_{ER}$ from Eq.~\ref{entropy_req}. Next, we meta-train the binary cross entropy and distillation loss $\mathcal{L}_{D}$ from Eq.~\ref{kd_eq}. At last, we meta-train with all loss functions together, where we weight $\mathcal{L}_{ER}$ with $\alpha = 0.01$ and $\mathcal{L}_{D}$ with $\beta=0.01$. We also noticed that the values of these two hyper-parameters depend on the complete loss function. As a result, we empirically set $\alpha = 0.1$ when the complete loss is composed of $\mathcal{L}_{ER}$ and $\mathcal{L}_{BCE}$; and keep $\beta=0.01$ when the complete loss is $\mathcal{L}_{D}$ and $\mathcal{L}_{BCE}$. We later analyze the sensitivity of the hyper-parameter in Sec.~\ref{Sec:ResDi}. Overall, the values have been found with grid search. A visual comparison of our objective is shown in Fig.~\ref{fig:ModelCOmponents}.

\begin{figure}[!htbp]
	\centering
	\begin{tabular}{cc}
	\subfigure[c][Average over data sets with FCRN.]{{\label{figWeidi_avg}\includegraphics[width=6cm]{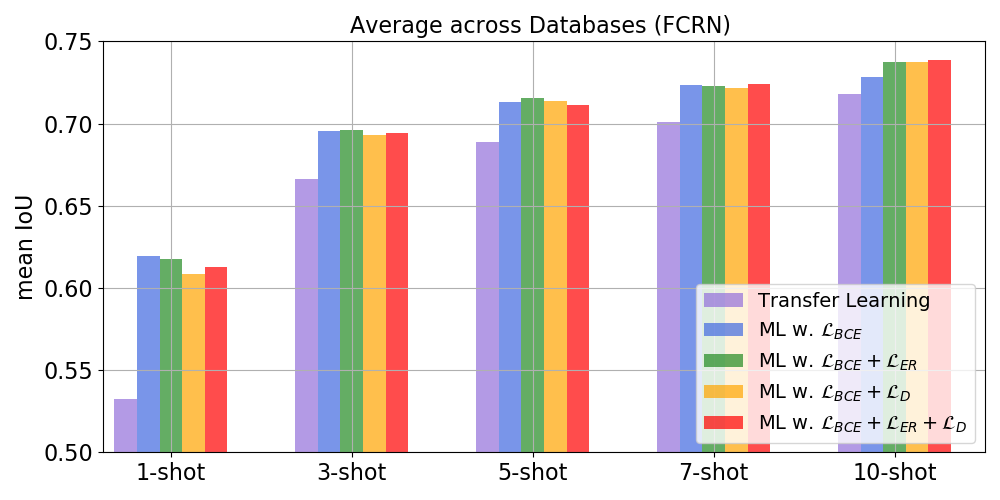}}}
	\subfigure[c][Average over data sets with U-Net.]{{\label{figUnet_avg}\includegraphics[width=6cm]{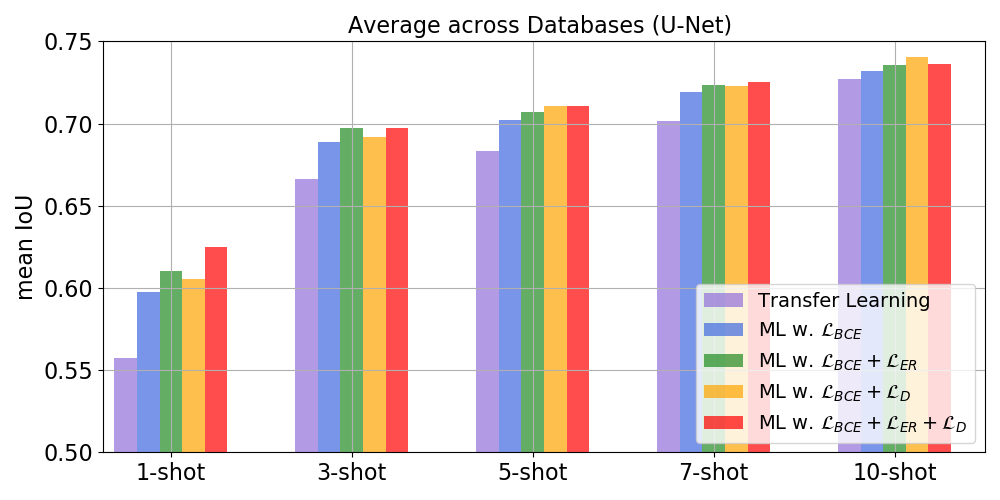}}}
    \end{tabular}
	\caption{Mean intersection over union (IoU) comparison using all datasets. We compare all loss function combinations and transfer learning for all data sets using the FCRN architecture in (a) and U-Net architecture in (b). }
	\label{figure_comp}
\end{figure}

\subsection{Results Discussion}\label{Sec:ResDi}

\begin{figure}[!htbp]

	\begin{tabular}{cc}
	\subfigure[c][TNBC, FCRN Net.]{{\label{figFCRN_TNBC}\includegraphics[width=6cm]{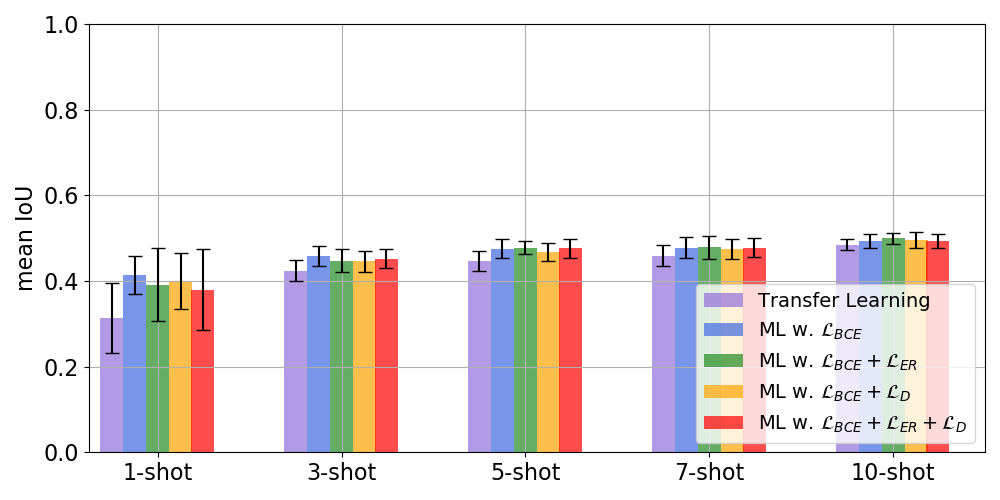}}}
	\subfigure[c][TNBC, U-Net.]{{\label{figUNet_TNBC}\includegraphics[width=6cm]{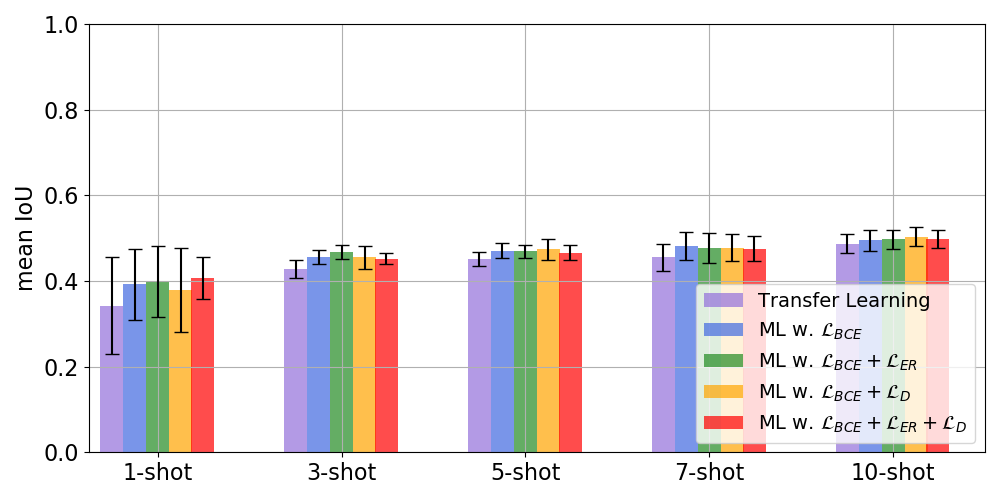}}}

    \end{tabular}
    \begin{tabular}{cc}
	\subfigure[c][EM, FCRN Net.]{{\label{figFCRN_EM}\includegraphics[width=6cm]{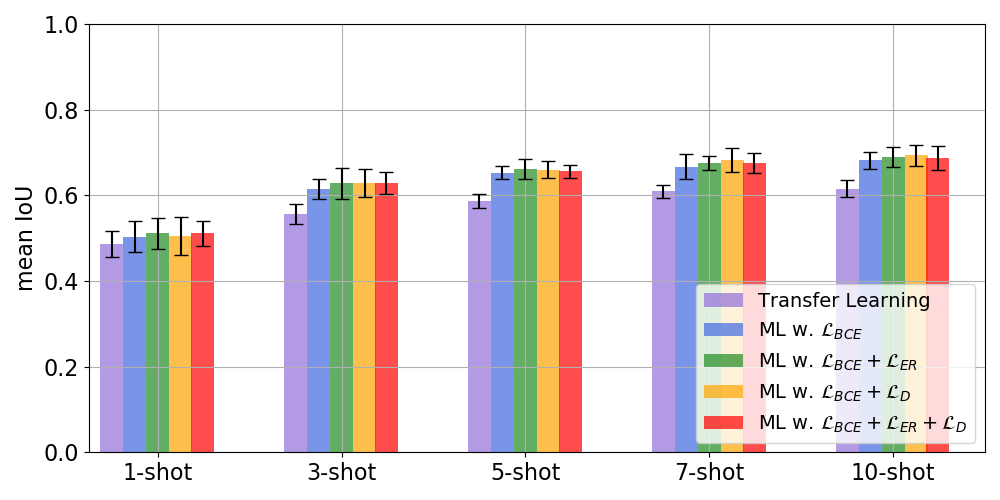}}}
	\subfigure[c][EM, U-Net.]{{\label{figUNet_EM}\includegraphics[width=6cm]{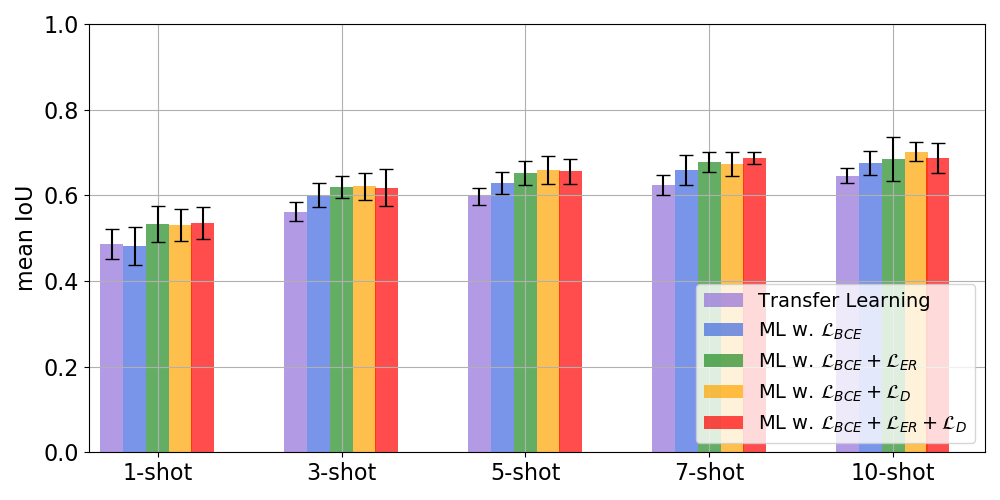}}}

    \end{tabular}
    \begin{tabular}{cc}
	\subfigure[c][ssTEM, FCRN Net.]{{\label{figFCRN_ssTEM}\includegraphics[width=6cm]{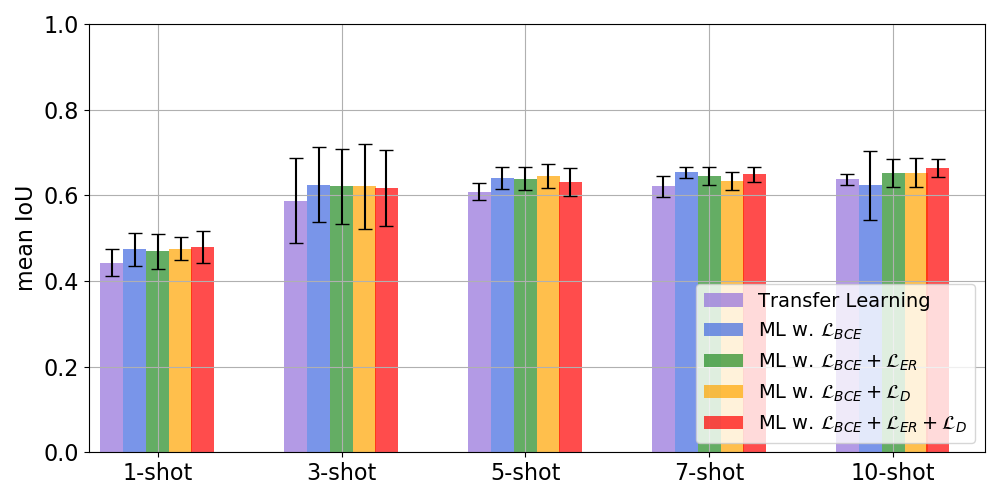}}}
	\subfigure[c][ssTEM, U-Net.]{{\label{figUNet_ssTEM}\includegraphics[width=6cm]{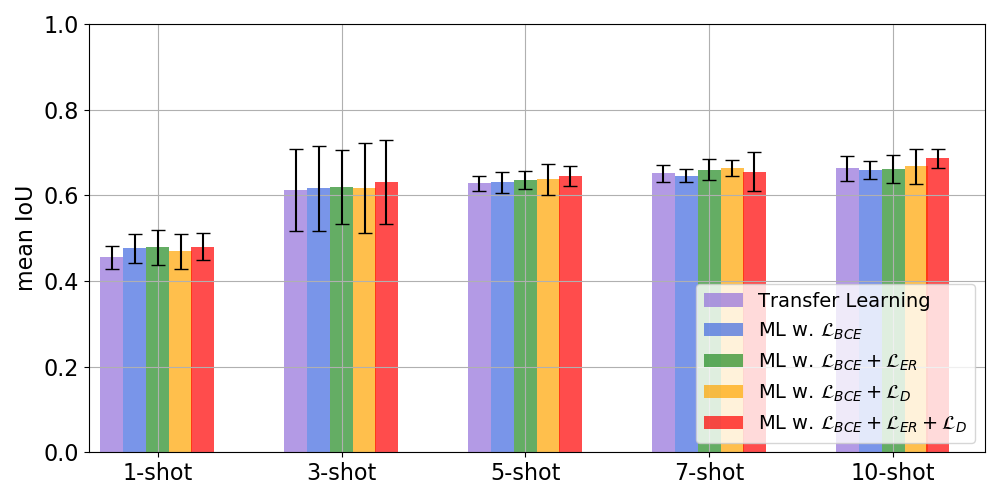}}}

    \end{tabular}
    \begin{tabular}{cc}
	\subfigure[c][B39, FCRN Net.]{{\label{figFCRN_B5}\includegraphics[width=6cm]{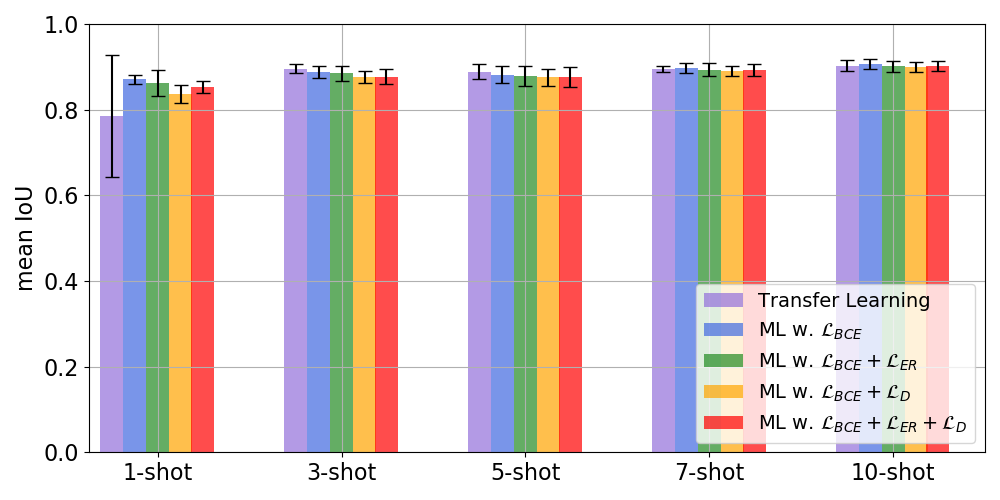}}}
	\subfigure[c][B39, U-Net.]{{\label{figUNet_B5}\includegraphics[width=6cm]{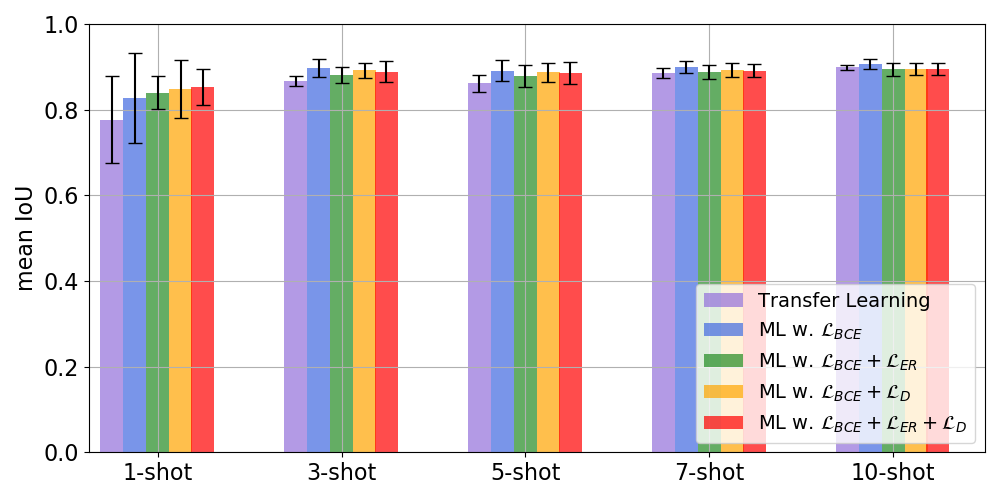}}}

    \end{tabular}
    \begin{tabular}{cc}
	\subfigure[c][B5, FCRN Net.]{{\label{figFCRN_B39}\includegraphics[width=6cm]{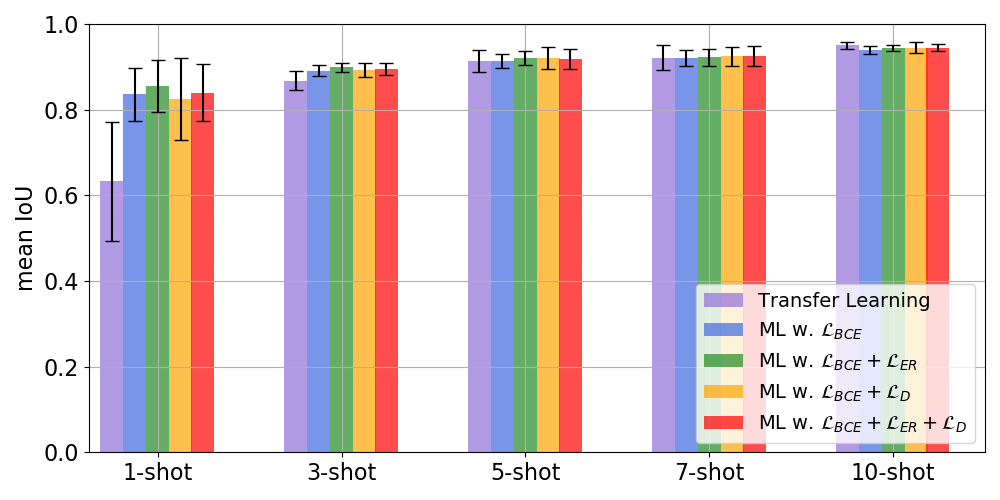}}}
	\subfigure[c][B5, U-Net.]{{\label{figUNet_B39}\includegraphics[width=6cm]{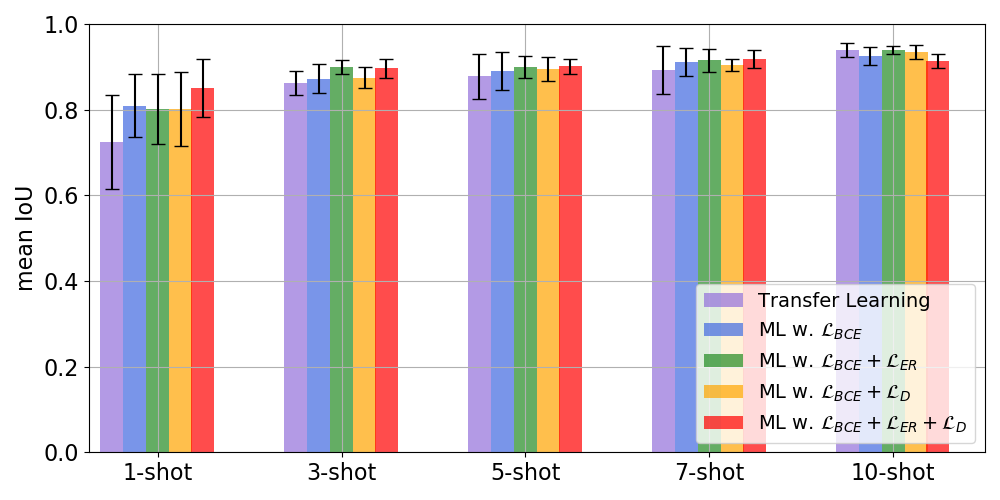}}}

    \end{tabular}
	\caption{Mean intersection over union (IoU) and standard deviation result comparison. We include of all loss functions and transfer learning the FCRN architecture (\textit{left} column) and U-Net architecture (\textit{right} column). ML stands for meta-learning.}
	\label{figure_comp_perdataset}
\end{figure}

We present the results of our approach with different combinations of the objective function as described in Sec.~\ref{protocol_eval}, as well as the complete model as presented in Algorithm~\ref{algo}. The quantitative evaluation is summarized in Fig.~\ref{figure_comp} and Fig.~\ref{figure_comp_perdataset}. In Fig.~\ref{figure_comp} we present the mean intersection over union (IoU) over all data sets while in  Fig.~\ref{figure_comp_perdataset} we present the mean IoU and the standard deviation over our ten random selections of the $K$-shot samples for each data set individually. We also provide a visual comparison with transfer learning in Fig.~\ref{fig:VisualComp1}.

At first, relying only on the binary cross entropy loss $\mathcal{L}_{BCE}$ for meta-training represents the baseline result for our approach. Adding the entropy regularization $\mathcal{L}_{ER}$ has a positive impact on some $K$-shot learning experiments. This can be seen in Fig.~\ref{figure_comp_perdataset}, which depicts meta-training with $\mathcal{L}_{BCE}$ and meta-training with $\mathcal{L}_{BCE} + \mathcal{L}_{ER}$. Similarly, the use of the distillation loss $\mathcal{L}_{D}$ together with the binary cross entropy, i.e. $\mathcal{L}_{BCE} + \mathcal{L}_{D}$ in Fig.~\ref{figure_comp_perdataset}, generally improves the mean IoU. The complete loss function $\mathcal{L}_{BCE} + \mathcal{L}_{ER} + \mathcal{L}_{D}$ gives the best results on average. This is easier to be noticed in Fig.~\ref{figWeidi_avg} and Fig.~\ref{figUnet_avg} where we compare the different objective combinations; and the results show that the complete loss produces the best results for most $K$ shot problems. In addition, it is clear (Fig.~\ref{figure_comp}) that our contributions, the $\mathcal{L}_{ER}$ and $\mathcal{L}_{D}$ objectives, have a positive impact on the outcome when more $K$-shot samples are added.

Comparing the FCRN with U-Net architectures in Fig.~\ref{figWeidi_avg} and Fig.~\ref{figUnet_avg}, we observe that U-Net delivers slightly better results for the complete loss combination. Nevertheless, the overall behavior is similar for both architectures. The standard binary cross entropy without our regularization and distillation objectives is already more accurate than transfer learning. Fig.~\ref{figure_comp_perdataset} also shows that the transfer learning performance can be comparable to ours (complete model) for 10-shot learning in some cases, e.g. Fig.~\ref{figFCRN_B5} and Fig.~\ref{figUNet_B5} for FCRN and U-Net. On 10-shot learning, transfer-learning benefits from the larger number of training data, resulting in performance closer to meta-learning. Other than the better performance, we also make two consistent observations when comparing to transfer learning. First, we argue that the main reason for the better performance is the parameter update rule. Transfer-learning averages gradients over shuffled samples, while meta-learning relies on task-based learning and then on averaging over the tasks. Second, meta-learning shows faster convergence. In particular,  meta-learning converges 40x faster for FCRN and around 60x faster for U-Net. This is an important advantage when working with multiple training sets. The disadvantage of meta-learning is the selection of the hyper-parameters. Our experience in the selection of the optimizer, including its hyper-parameters, as well as the selection of $\alpha$ and $\beta$ demonstrates that we need to be careful in this process. However, the optimizer and hyper-parameters are consistent for all tasks of cell segmentation. On the other hand, transfer learning involves only the selection of the optimizer. 

We can conclude the following points from the experiments. Meta-learning yields an overall stronger performance than transfer-learning as we average across our data sets, in addition, the new loss functions boosts the meta-learning performance in many cases, as shown in Figures ~\ref{figWeidi_avg} and ~\ref{figUnet_avg}. Both approaches, have low standard deviation when we add more shots in the fine-tuning step, as shown in Fig.~\ref{figure_comp_perdataset}. The same overall behavior is observed regardless the network architecture. Moreover, both architectures exhibit similar mean IoU. Besides, we notice that our regularization loss functions are not necessary for results around 90\% or more. In this case, only the additional shots make an impact. Finally, we notice the difference in performance across the data sets. For example, the mean IoU in TNBC target domain at 10-shot learning is within 50\% range while in EM we are able to reach around 70\% range. The reason could be that the source domains are more correlated with the EM target domain in terms of features than TNBC. Overall, our approach produces promising results when having at least 5-shots, i.e. annotated images, from the target domain for fine-tuning. This is a reasonably low amount of required to annotation to reach good generalization.

\section{Conclusion}\label{conclusion}
We have presented a few-shot meta-learning approach for microscopy image cell segmentation. The experiments show that our approach enables the learning of a generic and adaptable few-shot learning model from the available annotated images and cell segmentation problems of the source data sets. We fine-tune this model on the target data set using a limited annotated images. In the context of meta-training, we proposed the combination of three objectives to segment and regularize the predictions based on cross-domain tasks. Our experiments on five microscopy data sets show promising results from 1- to 10-shot meta-learning. As future work, we will include more data sets in our study to explore their correlation and impact on the target data set. Also, we plan to pose our problem in the context of domain generalization where the target data sets lacks of annotation. 

\section*{Acknowledgments}
This work was partially funded by Deutsche Forschungsgemeinschaft (DFG), Research Training Group GRK 2203: Micro- and nano-scale sensor technologies for the lung (PULMOSENS), and the Australian Research Council through grant FT190100525.  G.C. acknowledges the support by the Alexander von Humboldt-Stiftung for the renewed research stay sponsorship.

%
%

\begin{thebibliography}{10}
\providecommand{\url}[1]{\texttt{#1}}
\providecommand{\urlprefix}{URL }
\providecommand{\doi}[1]{https://doi.org/#1}

\bibitem{arteta2016counting}
Arteta, C., Lempitsky, V., Zisserman, A.: Counting in the wild. In: European
  conference on computer vision. pp. 483--498. Springer (2016)

\bibitem{belagiannis2018}
Belagiannis, V., Farshad, A., Galasso, F.: Adversarial network compression. In:
  Computer Vision -- ECCV 2018 Workshops. pp. 431--449. Springer International
  Publishing (2019)

\bibitem{de2015deep}
de~Brebisson, A., Montana, G.: Deep neural networks for anatomical brain
  segmentation. In: Proceedings of the IEEE conference on computer vision and
  pattern recognition workshops. pp. 20--28 (2015)

\bibitem{bronskill2020tasknorm}
Bronskill, J., Gordon, J., Requeima, J., Nowozin, S., Turner, R.E.: Tasknorm:
  Rethinking batch normalization for meta-learning. arXiv preprint
  arXiv:2003.03284  (2020)

\bibitem{chen2017deeplab}
Chen, L.C., Papandreou, G., Kokkinos, I., Murphy, K., Yuille, A.L.: Deeplab:
  Semantic image segmentation with deep convolutional nets, atrous convolution,
  and fully connected crfs. IEEE transactions on pattern analysis and machine
  intelligence  \textbf{40}(4),  834--848 (2017)

\bibitem{ciresan2012deep}
Ciresan, D., Giusti, A., Gambardella, L.M., Schmidhuber, J.: Deep neural
  networks segment neuronal membranes in electron microscopy images. In:
  Advances in neural information processing systems. pp. 2843--2851 (2012)

\bibitem{cirecsan2013mitosis}
Cire{\c{s}}an, D.C., Giusti, A., Gambardella, L.M., Schmidhuber, J.: Mitosis
  detection in breast cancer histology images with deep neural networks. In:
  International conference on medical image computing and computer-assisted
  intervention. pp. 411--418. Springer (2013)

\bibitem{clavera2018model}
Clavera, I., Rothfuss, J., Schulman, J., Fujita, Y., Asfour, T., Abbeel, P.:
  Model-based reinforcement learning via meta-policy optimization. arXiv
  preprint arXiv:1809.05214  (2018)

\bibitem{dhillon2019baseline}
Dhillon, G.S., Chaudhari, P., Ravichandran, A., Soatto, S.: A baseline for
  few-shot image classification. arXiv preprint arXiv:1909.02729  (2019)

\bibitem{dijkstra2018centroidnet}
Dijkstra, K., van~de Loosdrecht, J., Schomaker, L., Wiering, M.A.: Centroidnet:
  A deep neural network for joint object localization and counting. In: Joint
  European Conference on Machine Learning and Knowledge Discovery in Databases.
  pp. 585--601. Springer (2018)

\bibitem{dong2018few}
Dong, N., Xing, E.: Few-shot semantic segmentation with prototype learning. In:
  BMVC. vol.~3 (2018)

\bibitem{dou2019domain}
Dou, Q., de~Castro, D.C., Kamnitsas, K., Glocker, B.: Domain generalization via
  model-agnostic learning of semantic features. In: Advances in Neural
  Information Processing Systems. pp. 6447--6458 (2019)

\bibitem{faustino2009automatic}
Faustino, G.M., Gattass, M., Rehen, S., de~Lucena, C.J.: Automatic embryonic
  stem cells detection and counting method in fluorescence microscopy images.
  In: 2009 IEEE International Symposium on Biomedical Imaging: From Nano to
  Macro. pp. 799--802. IEEE (2009)

\bibitem{finn2017model}
Finn, C., Abbeel, P., Levine, S.: Model-agnostic meta-learning for fast
  adaptation of deep networks. In: Proceedings of the 34th International
  Conference on Machine Learning-Volume 70. pp. 1126--1135. JMLR. org (2017)

\bibitem{Gerhard2013}
Gerhard, S., Funke, J., Martel, J., Cardona, A., Fetter, R.: Segmented
  anisotropic sstem dataset of neural tissue. figshare pp.~0--0 (2013)

\bibitem{grandvalet2005semi}
Grandvalet, Y., Bengio, Y.: Semi-supervised learning by entropy minimization.
  In: Advances in neural information processing systems. pp. 529--536 (2005)

\bibitem{hinton2015distilling}
Hinton, G., Vinyals, O., Dean, J.: Distilling the knowledge in a neural
  network. arXiv preprint arXiv:1503.02531  (2015)

\bibitem{lehmussola2007computational}
Lehmussola, A., Ruusuvuori, P., Selinummi, J., Huttunen, H., Yli-Harja, O.:
  Computational framework for simulating fluorescence microscope images with
  cell populations. IEEE transactions on medical imaging  \textbf{26}(7),
  1010--1016 (2007)

\bibitem{li2018learning}
Li, D., Yang, Y., Song, Y.Z., Hospedales, T.M.: Learning to generalize:
  Meta-learning for domain generalization. In: Thirty-Second AAAI Conference on
  Artificial Intelligence (2018)


\bibitem{long2015fully}
Long, J., Shelhamer, E., Darrell, T.: Fully convolutional networks for semantic
  segmentation. In: Proceedings of the IEEE conference on computer vision and
  pattern recognition. pp. 3431--3440 (2015)

\bibitem{lu2015improved}
Lu, Z., Carneiro, G., Bradley, A.P.: An improved joint optimization of multiple
  level set functions for the segmentation of overlapping cervical cells. IEEE
  Transactions on Image Processing  \textbf{24}(4),  1261--1272 (2015)

\bibitem{lucchi2013learning}
Lucchi, A., Li, Y., Fua, P.: Learning for structured prediction using
  approximate subgradient descent with working sets. In: Proceedings of the
  IEEE Conference on Computer Vision and Pattern Recognition. pp. 1987--1994
  (2013)

\bibitem{mensink2012metric}
Mensink, T., Verbeek, J., Perronnin, F., Csurka, G.: Metric learning for large
  scale image classification: Generalizing to new classes at near-zero cost.
  In: European Conference on Computer Vision. pp. 488--501. Springer (2012)

\bibitem{munkhdalai2017meta}
Munkhdalai, T., Yu, H.: Meta networks. In: Proceedings of the 34th
  International Conference on Machine Learning-Volume 70. pp. 2554--2563. JMLR.
  org (2017)

\bibitem{naylor2017nuclei}
Naylor, P., La{\'e}, M., Reyal, F., Walter, T.: Nuclei segmentation in
  histopathology images using deep neural networks. In: 2017 IEEE 14th
  international symposium on biomedical imaging (ISBI 2017). pp. 933--936. IEEE
  (2017)

\bibitem{naylor2018segmentation}
Naylor, P., La{\'e}, M., Reyal, F., Walter, T.: Segmentation of nuclei in
  histopathology images by deep regression of the distance map. IEEE
  transactions on medical imaging  \textbf{38}(2),  448--459 (2018)

\bibitem{nichol2018first}
Nichol, A., Achiam, J., Schulman, J.: On first-order meta-learning algorithms.
  arXiv preprint arXiv:1803.02999  (2018)

\bibitem{ravi2016optimization}
Ravi, S., Larochelle, H.: Optimization as a model for few-shot learning  (2016)

\bibitem{rohrbach2013transfer}
Rohrbach, M., Ebert, S., Schiele, B.: Transfer learning in a transductive
  setting. In: Advances in neural information processing systems. pp. 46--54
  (2013)

\bibitem{ronneberger2015u}
Ronneberger, O., Fischer, P., Brox, T.: U-net: Convolutional networks for
  biomedical image segmentation. In: International Conference on Medical image
  computing and computer-assisted intervention. pp. 234--241. Springer (2015)

\bibitem{schmidhuber1992learning}
Schmidhuber, J.: Learning to control fast-weight memories: An alternative to
  dynamic recurrent networks. Neural Computation  \textbf{4}(1),  131--139
  (1992)

\bibitem{snell2017prototypical}
Snell, J., Swersky, K., Zemel, R.: Prototypical networks for few-shot learning.
  In: Advances in neural information processing systems. pp. 4077--4087 (2017)

\bibitem{tzeng2017adversarial}
Tzeng, E., Hoffman, J., Saenko, K., Darrell, T.: Adversarial discriminative
  domain adaptation. In: Proceedings of the IEEE Conference on Computer Vision
  and Pattern Recognition. pp. 7167--7176 (2017)

\bibitem{wahlby2004combining}
W{\"a}hlby, C., Sintorn, I.M., Erlandsson, F., Borgefors, G., Bengtsson, E.:
  Combining intensity, edge and shape information for 2d and 3d segmentation of
  cell nuclei in tissue sections. Journal of microscopy  \textbf{215}(1),
  67--76 (2004)
\bibitem{vinyals2016matching}
Vinyals, O., Blundell, C., Lillicrap, T., Wierstra, D., et~al.: Matching
  networks for one shot learning. In: Advances in neural information processing
  systems. pp. 3630--3638 (2016)

\bibitem{xie2018microscopy}
Xie, W., Noble, J.A., Zisserman, A.: Microscopy cell counting and detection
  with fully convolutional regression networks. Computer methods in
  biomechanics and biomedical engineering: Imaging \& Visualization
  \textbf{6}(3),  283--292 (2018)

\bibitem{xing2016robust}
Xing, F., Yang, L.: Robust nucleus/cell detection and segmentation in digital
  pathology and microscopy images: a comprehensive review. IEEE reviews in
  biomedical engineering  \textbf{9},  234--263 (2016)

\bibitem{zhang2012classifying}
Zhang, X., Wang, H., Collins, T.J., Luo, Z., Li, M.: Classifying stem cell
  differentiation images by information distance. In: Joint European Conference
  on Machine Learning and Knowledge Discovery in Databases. pp. 269--282.
  Springer (2012)

\end{thebibliography}

\end{document}